\documentclass[runningheads]{llncs}
\usepackage[T1]{fontenc}
\usepackage{algorithm}
\usepackage{algorithmic}
\usepackage{amsmath}
\usepackage{amsfonts}
\usepackage{mathtools}
\usepackage{color}
\usepackage{booktabs}
\usepackage{bm}
\usepackage{float}
\usepackage{subcaption}
\usepackage{caption}
\usepackage[pdftex]{graphicx}

\begin{document}
\title{Balancing Immediate Revenue and Future Off-Policy Evaluation in Coupon Allocation}
\titlerunning{Revenue and OPE Balance in Coupon Allocation}
\author{Naoki Nishimura\inst{1}\orcidID{0000-0002-6906-4323} \and
Ken Kobayashi\inst{2}\orcidID{0000-0002-6609-7488} \and
Kazuhide Nakata \inst{2}\orcidID{0000-0002-5479-100X}}
%
%
\authorrunning{N. Nishimura et al.}
\institute{Recruit. Co.,\, Ltd., 1-9-2 Marunouchi, Chiyoda-ku, Tokyo, Japan\\
\email{nishimura@r.recruit.co.jp}\\
 \and
Tokyo Institute of Technology, 2-12-1 Ookayama, Meguro-ku, Tokyo, Japan\\
\email{\{kobayashi.k.ar,nakata.k.ac\}@m.titech.ac.jp }}
\maketitle 
%
\begin{abstract}
Coupon allocation drives customer purchases and boosts revenue. 
However, it presents a fundamental trade-off between exploiting the current optimal policy to maximize immediate revenue and exploring alternative policies to collect data for future policy improvement via off-policy evaluation (OPE). 
To balance this trade-off, we propose a novel approach that combines a model-based revenue maximization policy and a randomized exploration policy for data collection. 
Our framework enables flexible adjustment of the mixture ratio between these two policies to optimize the balance between short-term revenue and future policy improvement. We formulate the problem of determining the optimal mixture ratio as multi-objective optimization, enabling quantitative evaluation of this trade-off. 
We empirically verified the effectiveness of the proposed mixed policy using synthetic data.
Our main contributions are: (1) Demonstrating a mixed policy combining deterministic and probabilistic policies, flexibly adjusting the data collection vs. revenue trade-off. (2) Formulating the optimal mixture ratio problem as multi-objective optimization, enabling quantitative evaluation of this trade-off.

\end{abstract}
\section{Introduction}
Marketing campaigns like discount coupon promotions are crucial for web service providers to encourage customer purchases and drive revenue growth~\cite{sigala2013framework,uehara2024robust,yang2022personalized}. 
Several studies have developed various coupon allocation algorithms~\cite{Li2020,Liu2023}, which are typically evaluated via online A/B testing. 
However, conducting online tests risks negatively impacting revenue due to inherent costs like coupon discounts~\cite{gilotte2018offline}.

To mitigate these risks, off-policy evaluation (OPE) has emerged as a crucial tool, allowing the evaluation of new policies using historical log data.
OPE has been a significant topic in reinforcement learning and causal inference for over two decades~\cite{precup2000eligibility}. 
Recent advancements have expanded OPE applications, as outlined by Uehara et al.~\cite{uehara2023off}. 
In coupon allocation, OPE plays a vital role in evaluating new strategies without costly online experiments~\cite{jiang2016doubly}.
However, deterministic allocation policies aimed at maximizing revenue can introduce bias, hindering OPE performance improvement~\cite{narita2022off}.

To address this issue, it is crucial to balance exploiting the current best policy for short-term revenue and exploring alternative actions to enable more accurate policy evaluation and improvement. 
This research aims to determine the optimal mixture ratio between model-based revenue maximization policies and random exploration policies for data collection. This balances short-term revenue goals with enabling accurate future OPE for policy improvement.

The main contributions of this research are twofold:
\begin{enumerate}
\item \textbf{Mixed Policy Application}: We applied a mixed policy combining model-based deterministic allocation and randomized exploration to the coupon allocation problem. This enables flexible adjustment of the trade-off between improving data collection efficiency for future OPEs and maximizing near-term revenue.
\item \textbf{Optimal Mixture Ratio Formulation}: We newly formulated the problem of determining the optimal mixture ratio $\boldsymbol{\alpha}$ that balances the aforementioned trade-off. By modeling it as a multi-objective optimization problem, we enabled quantitative evaluation and control over the balance between data collection efficiency and revenue acquisition.
\end{enumerate}
The effectiveness was empirically verified through experiments on synthetic data. By optimizing $\boldsymbol{\alpha}$, businesses can maximize expected revenue while ensuring reliable future OPE and policy improvement. This framework has potential applications beyond coupon allocation, in any context where the exploration-exploitation trade-off is relevant.

\section{Mixed Policy Framework for Revenue-OPE Balance}

Let $\boldsymbol{x}\in \mathcal{X}$ be the feature vector of users, $\boldsymbol{a}\in \mathcal{A}$ the action pattern of coupon allocation, and $r \in \mathbb{R}$ the observed reward when action $\boldsymbol{a}$ is taken for user $\boldsymbol{x}$, where $\mathcal{X}$ and $\mathcal{A}$ denote the feature and action spaces, respectively. We introduce the decision policy $\pi \colon \mathcal{X}\rightarrow \Delta(\mathcal{A})$ as a conditional probability distribution over the action space $\mathcal{A}$, with $\Delta(\mathcal{A})$ being the probability simplex over $\mathcal{A}$. Let the past data collection policies be $\pi_i(i=1,\ldots,K)$ and the new evaluation policy be $\pi_e$. The dataset collected by the data collection policy $\pi_i$ is denoted as $\mathcal{D}_i=\{(\boldsymbol{x}_j^{(i)},\boldsymbol{a}_j^{(i)},r_j^{(i)})\}_{j=1}^{n_i}$, where $n_i$ is the sample size of the dataset $\mathcal{D}_i$.

The naive policy evaluation estimator $\hat{V}_{\text{naive}}(\pi_{e})$~\cite{strehl2010learning} based on the dataset collected by the data collection policy $\pi_i$ is expressed as:
\begin{align}
\hat{V}_{\text{naive}}(\pi_{e}) = \frac{1}{ \sum_{j=1}^{n_i} \mathbb{I}\{h_{e}(\boldsymbol{x}_j^{(i)} ) = \boldsymbol{a}_j^{(i)} \}} \sum_{j=1}^{n_i} \mathbb{I}\{h_{e}(\boldsymbol{x}_j^{(i)} ) = \boldsymbol{a}_j^{(i)} \}r_j^{(i)}, \notag
\end{align}
where $\mathbb{I}\{\cdot\}$ denotes the indicator function and $h_e \colon \mathcal{X}\rightarrow \mathcal{A}$ is the action determined by the evaluation policy $\pi_{e}$ in the collected data. 
This estimator overestimates the value of actions more likely to be selected by the data collection policy. To address this bias, the inverse propensity score (IPS) estimator was introduced~\cite{dudik2011doubly}. Building on IPS, the balanced inverse propensity score (BIPS) estimator~\cite{kallus2018balanced} $\hat{V}_{\text{BIPS}}(\pi_{e})$ evaluates a policy $\pi_{e}$ using log data from multiple data collection policies:
\begin{align}
\hat{V}_{\text{BIPS}}(\pi_e) = \frac{1}{n}\sum_{i=1}^K \sum_{j=1}^{n_i} \frac{\pi_e(\boldsymbol{a}_j^{(i)}|\boldsymbol{x}_j^{(i)})}{\bar{\pi}_{1:K}(\boldsymbol{a}_j^{(i)}|\boldsymbol{x}_j^{(i)})}r_j^{(i)}, \notag
\end{align}
where $\bar{\pi}_{1:K}$ is the average policy defined by the weighted average of the $K$ different data collection policies with mixture ratios $\alpha_i$:
\begin{align}\label{eq:mixed-policy}
\bar{\pi}_{1:K}(\boldsymbol{a}_j^{(i)}|\boldsymbol{x}_j^{(i)}) = \sum_{i=1}^K \alpha_i \pi_{i}(\boldsymbol{a}_j^{(i)}|\boldsymbol{x}_j^{(i)}). \notag
\end{align}

\begin{figure}[htb]
\centering
\includegraphics[width=4.8in]{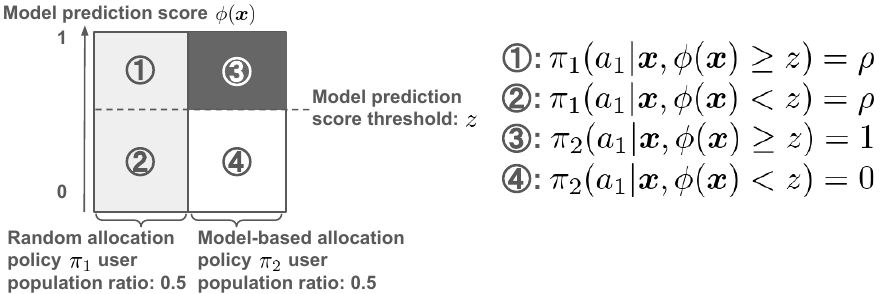}
\caption{Example of applying mixed data collection policies in coupon allocation using two policies for simplicity.}\label{c_example}
\end{figure}

To illustrate the concept of mixed data collection policies in coupon allocation, we consider a simplified example with two policies: a random allocation policy $\pi_1$ and a model-based allocation policy $\pi_2$, each applied with a 0.5 ratio (Fig. \ref{c_example}). While this example uses only two policies for clarity, the approach can be generalized to multiple policies.

Under $\pi_1$, coupon $a_1$ is provided with probability $\rho$, regardless of the model prediction score $\phi(\boldsymbol{x})$. Under $\pi_2$, users with $\phi(\boldsymbol{x}) \geq z$ receive a coupon with probability 1, and those with $\phi(\boldsymbol{x}) < z$ receive none. The probability of selecting $a_1$ under the average policy $\bar{\pi}_{1:2}$, with equal mixture ratios, is:
\begin{eqnarray}
p(a_1|\bar{\pi}_{1:2}(\boldsymbol{a}_j^{(i)}|\boldsymbol{x}_j^{(i)})) = \left\{
\begin{array}{ll}
0.5(\rho + 1) & (\phi(\boldsymbol{x}) \geq z)\\
0.5\rho & (\phi(\boldsymbol{x}) < z)
\end{array}
\right. \notag
\end{eqnarray}

When using the BIPS estimator, the weights for evaluating $a_1$ vary based on $\phi(\boldsymbol{x})$, with smaller weights when $\phi(\boldsymbol{x}) \geq z$ and larger when $\phi(\boldsymbol{x}) < z$. Mixing policies helps mitigate variance impact, especially when the model-based policy dominates.

To determine the optimal mixture ratio of data collection policies $\boldsymbol{\alpha}=(\alpha_1,...,\alpha_K)$, we formulate a multi-objective optimization problem balancing revenue and OPE performance:
\begin{align}
\min_{\boldsymbol{\alpha}}\quad &F(\boldsymbol{\alpha}) = \bigg(-f^r(\boldsymbol{\alpha}), f^e(\boldsymbol{\alpha}) \bigg)\notag \\
\text{s.t.}\quad &\sum_{i=1}^K \alpha_i = 1, \notag \\
& \boldsymbol{\alpha} \in [0, 1]^K. \notag 
\end{align}
The objective function represents the maximization of revenue (minimization of negative revenue) and the minimization of OPE error. By solving this optimization problem, we can obtain a Pareto optimal mixture ratio that maximizes the expected revenue while ensuring the performance of OPE.

It is important to note that our framework optimizes the mixture ratio of existing policies $\pi_i (i=1,\ldots,K)$ before new data collection. This approach balances revenue and OPE performance while avoiding risks associated with online testing of new policies. 
The resulting mixed policy is then used for data collection, enabling future OPE of new strategies.

\setcounter{footnote}{0}
\section{Experiments}
We conducted numerical experiments using synthetic data to verify the effectiveness of the proposed mixed policy approach. 
The code and detailed experimental setup are publicly available online\footnote{ https://github.com/nnnnishi/pricai2024balancing}.

\subsection{Experimental Setup}
We generated a synthetic dataset with 10,000 users, each with a four-dimensional feature vector sampled from a uniform distribution. 
The action space for coupon allocation was binary, indicating whether a coupon was allocated to a user or not. 
We designed three data collection policies: a random allocation policy using the first feature as the allocation probability, and two deterministic policies based on thresholds of the second and third features, respectively. 
The mixture ratios of these policies are denoted as $\alpha_1, \alpha_2, \alpha_3$. 
We also prepared two evaluation policies: one positively correlated and one negatively correlated with the second deterministic data collection policy.

The revenue metric $f^r(\boldsymbol{\alpha})$ is the true expected revenue under the mixed policy, while the OPE performance metric $f^e(\boldsymbol{\alpha})$ is the MSE between the true policy value and the BIPS estimate. 
We used Optuna v3.6.0~\cite{akiba2019optuna} with the NSGA-II algorithm for multi-objective optimization, setting the number of trials to 1,000.

\subsection{Results and Conclusion}
\begin{figure}[htb]
  \begin{minipage}[b]{0.48\hsize}
    \centering
    \includegraphics[width=2.3in]{{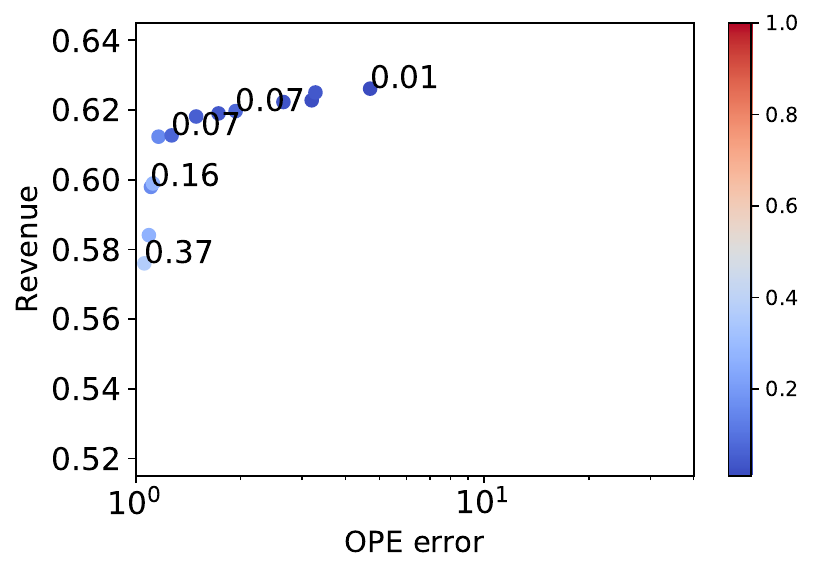}}
    \subcaption{Evaluation policy positively \\correlated with data collection policies}\label{curve_syn1}
  \end{minipage}
  \begin{minipage}[b]{0.48\hsize}
    \centering
    \includegraphics[width=2.3in]{{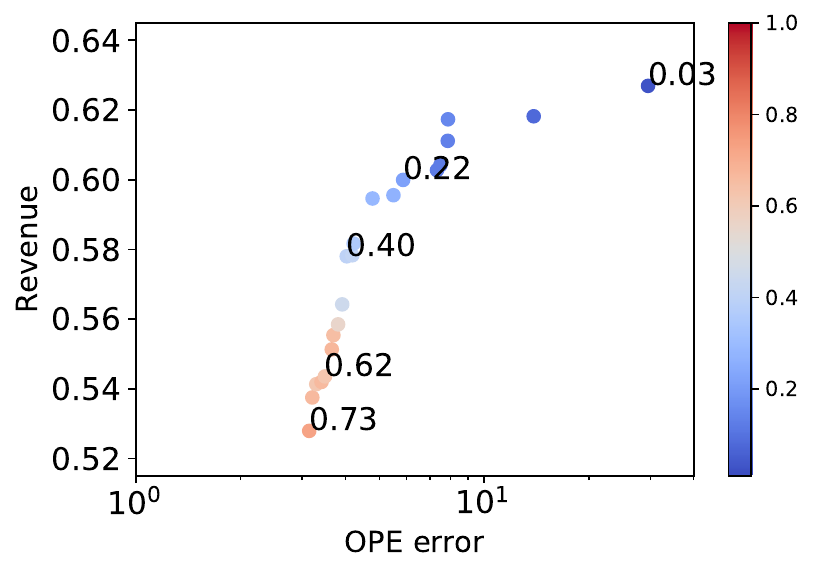}}
    \subcaption{Evaluation policy negatively \\correlated with data collection policies}\label{curve_syn2}
  \end{minipage}
  \caption{Pareto frontier of revenue and OPE error trade-off for different policy mixture ratios. Heatmap and labels indicate random assignment ratio.}\label{fig:curve_syn}
\end{figure}

Fig. \ref{fig:curve_syn} presents the Pareto frontier between revenue and OPE performance metrics based on the mixture ratio $\boldsymbol{\alpha}$ for both evaluation policies.

The results reveal key insights on revenue and OPE performance. 
As the random allocation policy ratio decreases, the revenue metric improves for both evaluation policies, but OPE performance deteriorates. 
OPE performance improves when probability distributions between data collection and evaluation policies are closer. 
The trade-off balance shows that reducing the random assignment ratio leads to higher revenue metrics but can negatively impact OPE performance, especially when the evaluation policy's distribution is close to deterministic data collection policies. 
The Pareto frontiers demonstrate policy mixture effectiveness, showing mixed policies can achieve better trade-offs compared to single policies in coupon allocation.

These experiments validate our proposed mixed policy approach in balancing revenue maximization and reliable OPE. 
Our study's novelty lies in its simultaneous consideration of OPE performance and revenue, extending beyond the single-objective focus of previous research. By quantitatively optimizing $\boldsymbol{\alpha}$, businesses can maximize expected revenue while ensuring reliable future OPE and policy improvement.

While using a simplified synthetic dataset, the framework is designed for complex real-world scenarios. 
Our approach provides insights into potential A/B test outcomes without online testing risks.
Future work could explore diverse feature distributions and intricate revenue functions. Comparing our method with actual A/B testing in real-world domains and incorporating advanced OPE methods~\cite{farajtabar2018more,kiyohara2023towards} could provide valuable insights. 

\begin{credits}
\subsubsection{\discintname}
The authors have no competing interests to declare that are
relevant to the content of this article. 
\end{credits}
\vspace{-1em}
%
%
%
\bibliographystyle{splncs04}
\bibliography{mybibliography}

\end{document}